%% file: access.tex
\documentclass{ieeeaccess}
\usepackage{cite}
\usepackage{amsmath,amssymb,amsfonts}
\usepackage{algorithmic}

\usepackage{graphicx}
\graphicspath{{Images/}}
\usepackage{float}
\usepackage[caption=false,font=footnotesize,labelfont=sf,textfont=sf]{subfig}

\usepackage{mathtools}
\DeclarePairedDelimiter{\abs}{\lvert}{\rvert}
\usepackage{amsthm}  %
\usepackage{thmtools}  %

\newcommand{\argmin}[1]{\underset{#1}{\operatorname{arg}\,\operatorname{min}}\;}

\theoremstyle{plain}  %
\newtheorem{proposition}{Proposition}  %

\declaretheoremstyle[%
	spaceabove=-6pt,%
    spacebelow=6pt,%
    headfont=\normalfont\itshape,%
    postheadspace=1em,%
    qed=\qedsymbol%
]{mystyle}

\DeclarePairedDelimiterX{\norm}[1]{\lVert}{\rVert}{#1}

\usepackage{caption,setspace}
\usepackage{textcomp}
\def\BibTeX{{\rm B\kern-.05em{\sc i\kern-.025em b}\kern-.08em
    T\kern-.1667em\lower.7ex\hbox{E}\kern-.125emX}}

\begin{document}
\history{Date of publication xxxx 00, 0000, date of current version xxxx 00, 0000.}
\doi{10.1109/ACCESS.2017.DOI}

\title{TSViz: Demystification of Deep Learning Models for Time-Series Analysis}

\author{\uppercase{Shoaib Ahmed Siddiqui}\authorrefmark{1}\authorrefmark{2}*, \uppercase{Dominique Mercier}\authorrefmark{1}\authorrefmark{2}*, \uppercase{Mohsin Munir}\authorrefmark{1}\authorrefmark{2}, \uppercase{Andreas Dengel}\authorrefmark{1}\authorrefmark{2}, and \uppercase{Sheraz Ahmed}\authorrefmark{1}.}
\address[1]{German Research Center for Artificial Intelligence (DFKI), Kaiserslautern, Germany.}
\address[2]{TU Kaiserslautern, Kaiserslautern, Germany.}

\tfootnote{*Equal contribution \\ This work was supported by the BMBF project DeFuseNN (Grant 01IW17002) and the NVIDIA AI Lab (NVAIL) program.}

\markboth
{Siddiqui \headeretal: TSViz - Demystification of Deep Learning Models for Time-Series Analysis}
{Siddiqui \headeretal: TSViz - Demystification of Deep Learning Models for Time-Series Analysis}

\corresp{Corresponding author: Shoaib Ahmed Siddiqui (E-mail: shoaib\_ahmed.siddiqui@dfki.de).}

\begin{abstract}
This paper presents a novel framework for demystification of convolutional deep learning models for time-series analysis.  This is a step towards making informed/explainable decisions in the domain of time-series, powered by deep learning. There have been numerous efforts to increase the interpretability of image-centric deep neural network models, where the learned features are more intuitive to visualize. 
Visualization in time-series domain is much more complicated as there is no direct interpretation of the filters and inputs as compared to the image modality. In addition, little or no concentration has been devoted for the development of such tools in the domain of time-series in the past. 
TSViz provides possibilities to explore and analyze a network from different dimensions at different levels of abstraction which includes identification of parts of the input that were responsible for a prediction (including per filter saliency), importance of different filters present in the network for a particular prediction, notion of diversity present in the network through filter clustering, understanding of the main sources of variation learnt by the network through inverse optimization, and analysis of the network's robustness against adversarial noise.
As a sanity check for the computed influence values, we demonstrate results regarding pruning of neural networks based on the computed influence information.
These representations allow to understand the network features so that the acceptability of deep networks for time-series data can be enhanced. This is extremely important in domains like finance, industry 4.0, self-driving cars, health-care, counter-terrorism etc., where reasons for reaching a particular prediction are equally important as the prediction itself. 
We assess the proposed framework for interpretability with a set of desirable properties essential for any method in this direction.
\end{abstract}


\maketitle

\input{Introduction}

\input{Related}

\input{Datasets}

\input{Approach}

\input{Evaluation}

\input{TSViz_Tool} %

\input{Discussion}

\input{Conclusion}

\bibliography{IEEEabrv.bib,bibliography.bib}
\bibliographystyle{IEEEtran}

\EOD

\end{document}

%% file: Introduction.tex
\section{Introduction} \label{sec:introduction}

Despite of astonishing results from deep learning based models in a range of applications which includes computer vision, speech analysis, translation systems etc.~\cite{NIPS2012_AlexNet,NIPS2010_Phone_Recognition,google_nmt}, there has been limited applicability of these high performance models due to their black-box nature and lack of explainability of their decisions~\cite{senn2018}. This is specifically applicable for domains like business, finance, natural disaster management, health-care, self-driving cars, industry 4.0 and counter-terrorism where reasons of reaching a particular decision are equally important as the prediction itself~\cite{mitTechReview2017}. 

There have been significant attempts to uncover the black-box nature of these deep learning based models~\cite{yosinski-2015-ICML-DL-understanding-neural-networks,DBLP:journals/corr/ZeilerF13,DBLP:journals/corr/SimonyanVZ13,timeseries-viz,informationBottleneckTheory,understandingDLrequiresRethinkingGeneralization}, where visualization of the model has been the most common strategy. 
Almost all of the proposed visualization systems are image-centric where visualizing the images is directly interpretable for humans (natural association to similar looking objects like eyes, faces, dogs, cars etc.). 
Most of the ideas are equally applicable to time-series, but unintuitive nature of the time-series data makes it difficult to directly transfer these ideas for improved human understanding of the model.

This paper presents a novel framework for demystification of deep models for time-series analysis. In particular, the contributions of this paper are manifold:
\begin{itemize}

\item An influence tracing algorithm to compute the input saliency map, which enables an understanding of the regions of the input that were responsible for a particular prediction.

\item An approach to compute the filter's influence using the proposed influence tracing algorithm. Filter importance is computed based on its influence on the final output. This information provides an idea to the user regarding the filters of the network that were important for a particular prediction.

\item An approach to discover the diversity present in the network based on filter clustering. Filters belonging to the same cluster exhibits similar behavior in terms of their activation pattern, therefore, responding to the same concept/feature.

\item An evaluation of pruning of the network leveraging the influence information as a sanity check for the utility of the information encapsulated into the computed influence.
The aim of this evaluation is to move towards principled design of network where the complexity of the problem aligns well with the complexity of the network. 
With the computed influences, it is possible to identify parts of the network responding/tuned to highly specific stimulus. These parts which contribute to the overfitting of the network can be pruned to promote generalization.

\item An inverse optimization framework where we optimize the input considering the network parameters as fixed.
This inverse optimization based framework enables the user to understand the main sources of variation learnt by the network in the input space.

\item An approach for demystification of the network through white-box gradient based adversarial attack methods (FGSM~\cite{FGSM} and iterative FGSM~\cite{iterativeFGSM}) which provides an understanding of the impact of adversarial noise to the inspected model. 
To the best of our knowledge, this is the first attempt to understand the impact of adversarial noise on deep learning models for time-series analysis.
This evaluation helps in answering two different questions: (a) Robustness of the network against adversarial noise and (b) network's understanding of the changes in input which can bring maximal changes to the output. 
The second question, albeit being more interesting one from our perspective, is sometimes difficult to answer in cases where the network is highly susceptible to adversarial noise.

\item A novel 3D visualization framework for time-series deep learning models. This framework is generic and capable of visualizing any convolutional deep learning model for time-series analysis. 
TSViz provides an opportunity to explore the network at different levels of abstraction i.e. from abstract to detailed view. The details regarding the user-interface are provided in Section~\ref{sec:implementation}.

\end{itemize}

This paper is structured as follows. We provide a brief overview of the previous work in the direction of demystification of deep learning models. We then provide details regarding the datasets and the network architectures that we employed in our experimental setup. The presented approach then covers both the method and the related experimentation. We then present an assessment of the proposed method based on a set of desirable properties essential for any method related to network interpretability. 
Finally we present the implementation details and provide some discussion regarding the obtained results before concluding the paper.

%% file: Related.tex
\section{Related Work} \label{sec:related}

Significant efforts have been made to understand the learnt features of a model in order to demystify the deep learning black-box. This is specifically the case for visual recognition models since visualizing the kernels themselves and their activations can give hints about the feature that the system is learning and the kernels themselves are directly interpretable by humans~\cite{yosinski-2015-ICML-DL-understanding-neural-networks,DBLP:journals/corr/ZeilerF13,DBLP:journals/corr/SimonyanVZ13,LRP_Bach_15}. 
One of the most common and effective technique for network visualization (specifically for visual modality) is saliency maps~\cite{nvidia-steeringAngle} which highlights the regions which the network focused on in order to generate a particular prediction. 
The Layer-wise Relevance Propagation (LRP) framework presented a different approach for tracing back the influence using relevance instead of the gradients~\cite{LRP_Bach_15}.
In order to visualize human understandable features/concept that the neuron is responding to, there have been significant efforts in detecting which input maximally activates a neuron~\cite{DBLP:journals/corr/GirshickDDM13}. The problem of discovering the part of input that was most responsible for a particular prediction has also been extensively studied~\cite{yosinski-2015-ICML-DL-understanding-neural-networks,DBLP:journals/corr/ZeilerF13,DBLP:journals/corr/SimonyanVZ13}. Shrikumar et al. (2017)~\cite{alternateToGradientBasedMaximalActivations} presented an alternate formulation from gradient based methods for obtaining the maximally activating input. Li et al. (2016)~\cite{textSaliencyMaps} used saliency maps to identify focus regions for textual data.

Despite of these advancements, the area of network visualizations for time-series analysis has remained unexplored till now. A recent attempt has been made by Kumar et al. (2017)~\cite{timeseries-viz} to visualize the input points which were most influential for a particular prediction through gradients (saliency).

Theoretical contributions have also been made in order to understand the amazing generalization capabilities of these deep models. Zhang et al. (2016)~\cite{understandingDLrequiresRethinkingGeneralization} presented empirical analysis to divert attention to the philosophical topic of what actually is perceived as generalization. Information bottleneck theory~\cite{informationBottleneckTheory} has also gained popularity as method for explaining the generalization and learning capabilities of deep learning based models. Montavon et al. (2017)~\cite{understandingDLPredviaInfluenceFunctions} presented influence functions as a methodology to trace back model predictions in terms of its training data. This analysis enabled discovery of dataset errors, model debugging, and creation of visually-indistinguishable adversarial training examples which are able to flip the network's test time predictions.

We refer the readers to~\cite{visualizationReviewICASSP2017} for a comprehensive review of the previous work in this direction based on the tutorial given at ICASSP (2017).

%% file: Datasets.tex
\section{Datasets and Architectures} \label{sec:datasets}

We opted for two different problem formulations common in time-series regime: Time-series regression and time-series classification.

\subsection{Regression}

For the regression setting, we trained a network on the Internet Traffic Dataset~\cite{cortez2012multi} for time-series forecasting (we only used the B5M dataset). The network operates on an input size of 50 time-steps and is comprised of three convolutional, two max-pooling and one dense layer. The network is trained using Adam optimizer with a learning rate of 0.001 for 5000 epochs using a batch size of 5 and Mean Squared Error (MSE) as the loss function. This network was able to achieve state-of-the-art forecasting results on the aforementioned dataset. The network takes in a bi-channel input where the first channel corresponds to the original input signal while the second channel is comprised of the first-order derivative of the original signal.

\subsection{Classification}

For the classification setting, we created a dummy time-series anomaly detection dataset for binary classification. The dataset is comprised of 60000 sequences of 50 time-steps each where each time-step contains values for pressure, temperature and torque. We randomly introduce point anomalies in the dataset and mark the sequences containing such point anomalies as anomalous. We never introduced any anomalies in the pressure signal. 
The dataset is split into 50000 train sequences and 10000 test sequences. 
The network is comprised of three convolutional layers comprising of 16, 32 and 64 filters respectively followed by a single dense layer. The network is trained using binary cross-entropy loss. %
The hyper-parameters were chosen based on our best guess without any cross-validation since the focus of our work is on interpretability instead of performance.

%% file: Approach.tex
\section{TSViz: The Proposed Approach} \label{sec:approach}

TSViz provides the possibility to interpret any convolutional network from several dimensions, at different levels of abstraction. This includes the global picture like the types and ordering of different layers present in the network, and their corresponding number of filters, moving onto more detailed information like parts of the input that each filter is responding to (Section~\ref{sec:input_influence}) as well as their importance (Section~\ref{sec:filter_influence}). This also includes filter grouping where filters which are exhibiting similar behavior are clustered together (Section~\ref{sec:cluster}), which captures the notion of network diversity.
TSViz also uncovers other hidden aspects of the network based on inverse optimization (Section~\ref{sec:invOptim}) and adversarial examples (Section~\ref{sec:advEx}).

\subsection{Background}\label{sec:backprop}

TSViz is based on the principle of backpropagation proposed by Rumelhart et al. (1986)~\cite{rumelhart1986learning}, which is essentially the chain rule from calculus.
Backpropagation algorithm provides an efficient way to compute influences of the tensor w.r.t. another tensor.
The same framework is the workhorse for training of the deep learning models where the influence of the network parameters is computed on the final cost/loss function. We leverage this capability to compute influences for uncovering the deep learning black-box by computing the influence of the input on the current filter which is the input saliency map. This also enables the discovery of the filter's importance by computing its influence on the final prediction of the network.
Therefore, this section provides a short recapitulation of the basic concept of the backpropagation algorithm along with laying out the necessary notation to be used later. 

The basic aim of learning in neural network is to reduce the loss function $\mathcal{L}: \mathbb{R} \times \mathbb{R} \mapsto \mathbb{R}_{+}$ ($\mathcal{L}: \mathbb{R}^C \times \mathbb{R}^C \mapsto \mathbb{R}_{+}$ in case of multi-class classification where C denotes the number of classes). The loss function captures the discrepancy between the network's prediction and the desired output. Ideally, the network must output a value which is same as the target. The whole learning process is about reducing the discrepancy between the two values. As the network output is calculated based on the weights and biases of the different neurons involved, these weights and biases are adapted during the learning process in order to reduce the loss function. Ultimately, backpropagation is about understanding how the change in the weights and biases of a network affect it's loss and to update the network parameters in the direction with the maximum decrease in the loss function. This computation of the optimal direction can be obtained by calculating the partial derivatives of the loss function with respect to any weight $W$ or bias $b$ as $\partial \mathcal{L}/\partial W$ and $\partial \mathcal{L}/\partial b$. 

Backpropagation algorithm can be decomposed into four steps which includes:
(i) Feed-forward pass through the network,
(ii) Backpropagation to the output layer,
(iii) Backpropagation to the hidden layers,
(iv) Updating network parameters.
In the feed-forward pass through the network, the output of all the hidden neurons is computed which is then used for the computation of the final network output. This evaluation is based on the randomly initialized weights. Based on the computed output, the final loss function is evaluated. The networks in deep learning are mostly comprised of both convolutional and dense layers. Rectified Linear Unit (ReLU) or some variant of it is commonly used as the non-linearity/activation function. The activation for the dense and convolutional layers is presented in Eq.~\ref{eq:fp_fc2} and Eq.~\ref{eq:fp_conv2} respectively. $a_{j}^{l}$ denotes the activation of the j\textsuperscript{th} neuron in the l\textsuperscript{th} layer (for dense layers) while $a_{ji}^{l}$ denotes the activation of the j\textsuperscript{th} neuron in the l\textsuperscript{th} layer at the i\textsuperscript{th} input location (for convolutional layers). $k$ is defined as $\lfloor Filter Size / 2 \rfloor$ for convolutional layers while $|z_j^{l-1}|$ denotes the number of neurons in the previous layer $l-1$.

\begin{equation} \label{eq:fp_fc1}
    z_{j}^{l} = \sum_{k=1}^{|z_j^{l-1}|} W_{jk}^{l} a_{k}^{l-1} + b_{j}^{l} %
\end{equation}

\begin{equation} \label{eq:fp_fc2}
	a_{j}^{l} = max\left(z_{j}^{l}, 0\right) %
\end{equation}

\begin{equation} \label{eq:fp_conv1}
	z_{ji}^{l} = \sum_{-k}^{k} W_{jk}^{l} a_{i-k}^{l-1} + b_{j}^{l}
\end{equation}

\begin{equation} \label{eq:fp_conv2}
	a_{ji}^{l} = max\left(z_{ji}^{l}, 0\right)
\end{equation}

The error is backpropagated to the initial layers, and the gradient with respect to the network parameters is computed (weights and biases). The error $\delta$ of $j^{th}$ neuron at the output layer $L$ is presented in Eq.~\ref{eq:bp2}.

\begin{equation} \label{eq:bp1}
	\delta_{j}^{L} = \frac{\partial \mathcal{L}}{\partial a_{j}^{L}} \frac{\partial a_{j}^{L}}{\partial z_{j}^{L}}
\end{equation}

\begin{equation} \label{eq:bp2}
	\delta_{j}^{L} = \frac{\partial \mathcal{L}}{\partial a_{j}^{L}} {max}'\left(z_{j}^{L}, 0\right)
\end{equation}

\begin{equation} \label{eq:bp3}
	\delta_{j}^{l} = \left(\left(W_{ij}^{l}\right)^T \delta_{j}^{l+1} \right) \odot max'\left(z_{j}^{l}, 0\right)
\end{equation}

\begin{equation} \label{eq:bp4}
max'(x, 0)=
\begin{cases}
1, & \text{if}\ x > 0 \\
0, & \text{otherwise}
\end{cases}
\end{equation}

Once gradient of the loss w.r.t. output layer is computed, the error is backpropagated to all neurons in the hidden layers using Eq.~\ref{eq:bp3}. The gradient of ReLU is 1 where the value of input $x$ exceeds 0 and remains 0 otherwise as mentioned in Eq.~\ref{eq:bp4}. Similarly, the max-pooling layer has gradient equal to 1 wherever the maximum quantity occurred and remains 0 otherwise. The rate of change of loss $\mathcal{L}$ w.r.t. the bias and weights in the l\textsuperscript{th} layer is given in Eq.~\ref{eq:bp5} and Eq.~\ref{eq:bp6} respectively.

\begin{equation} \label{eq:bp5}
	\frac{\partial \mathcal{L}}{\partial b_{j}^{l}} = \delta_{j}^{l}
\end{equation}

\begin{equation} \label{eq:bp6}
	\frac{\partial \mathcal{L}}{\partial W_{jk}^{l}} = a_{k}^{l-1}\delta_{j}^{l}
\end{equation}

After computation of the gradients, the network parameters (weights and biases) are updated in the negative gradient direction as this is the optimal direction for maximum reduction in the loss.

\subsection{Influence Computation}\label{sec:influence}

TSViz contributes to the interpretability of deep learning models designed specifically for time-series analysis tasks at different levels of abstraction. The first and one of the most intuitive explanation for any model is based on the influence of the input on the final prediction. Consequently, this influence can also be computed for the intermediate states of the network (both in the forward as well as the backward direction). 

There are two different influences that can be computed based on any particular filter in the network. The first influence stems from the fact that the input has an impact on the outputs of the particular filter in question (Section~\ref{sec:input_influence}) while the second influence is of the filter itself on the final output (Section~\ref{sec:filter_influence}). We will now visit each of these influences in detail.

\subsubsection{INPUT INFLUENCE}\label{sec:input_influence}

\begin{figure*}[h!]
\centering
\subfloat[Input]{\includegraphics[width=0.44\linewidth]{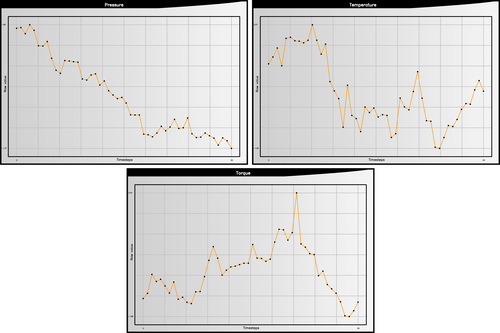}
\label{fig:input_raw}}
\hfil
\hfil
\hfil
\subfloat[Input with importance]{\includegraphics[width=0.44\linewidth]{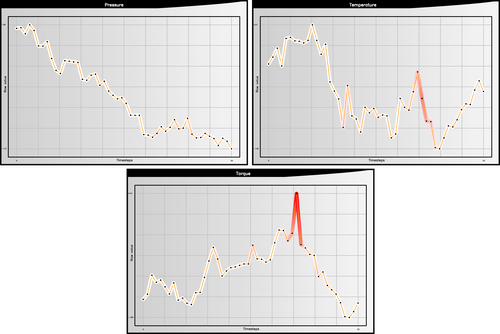}
\label{fig:input_sal}}
\hfil
\caption{A particular anomalous example provided to the network (Anomaly detection dataset)}
\label{fig:input}
\end{figure*}

The first influence originates from the input.
This information provides important insights regarding the data points in the input that the network is actually responding to for computation of its output. The information regarding the parts of the input which were responsible for a particular prediction is considered a viable explanation in many scenarios including domains like self-driving cars~\cite{textualExplanationsForSelfDrivingCars}, finance~\cite{timeseries-viz} and medical imaging~\cite{mriDataPredictiveDifferenceAnalysis}.
It is important to note that we also compute the input's influence for every filter along with the final output.

This value can be obtained by computing the gradient of the current layer $l$ w.r.t. the input layer. 
We use the absolute value of the gradient as the magnitude is of relevance irrespective of the direction. We denote the input as $a^{0}$, therefore, this influence of the input can be computed using Eq.~\ref{eq:inputInfluence3}.

\begin{equation} \label{eq:inputInfluence1}
\delta_{j}^{l} = \frac{\partial a^{l}}{\partial z_{j}^{1}} \frac{\partial z_{j}^{1}}{\partial a_{j}^{0}}
\end{equation}

\begin{equation} \label{eq:inputInfluence2}
\delta_{j}^{0} = \frac{\partial a^{l}}{\partial a_{j}^{0}}
\end{equation}

\begin{equation} \label{eq:inputInfluence3}
    I_{input}^{d} = \abs{\delta^{0}}
\end{equation}

In order to be able to visualize and compare the saliency of the different filters, the absolute values of the influences are scaled using the min-max scaling presented in Eq.~\ref{eq:minMaxSc}.

\begin{equation} \label{eq:minMaxSc}
	I_{input}^{l} = \frac{I_{input}^{l} - \min I_{input}^{l}} {\max I_{input}^{l} - \min I_{input}^{l}}
\end{equation}

Fig.~\ref{fig:input_raw} visualizes a sample of an anomalous input in the anomaly detection dataset. Fig.~\ref{fig:input_sal} equips the raw filter output with the saliency information to provide a direct interpretation of its utility. It is evident from the figure that the network focused on sudden peaks present in the input to mark the sequence as anomalous.
This saliency is computed using Eq.~\ref{eq:inputInfluence3} after applying min-max normalization. 

\subsubsection{FILTER INFLUENCE}\label{sec:filter_influence}

\begin{figure*}[h!]
\centering
\subfloat[Filters]{\includegraphics[width=0.33\linewidth]{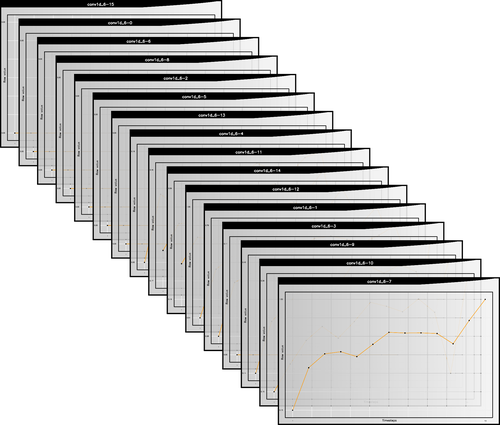}
\label{fig:conv3_raw}}
\subfloat[Filters with importance and saliency]{\includegraphics[width=0.33\linewidth]{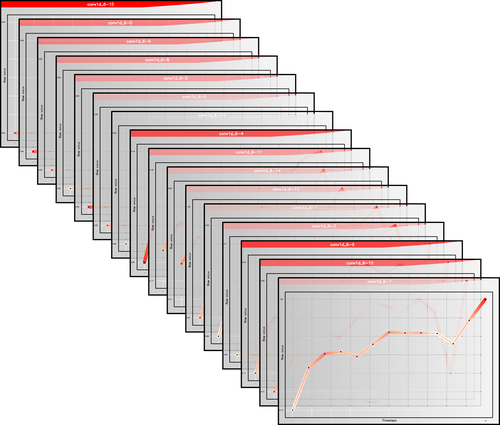}
\label{fig:conv3_imp}}
\subfloat[Filters with importance, saliency and clusters]{\includegraphics[width=0.34\linewidth]{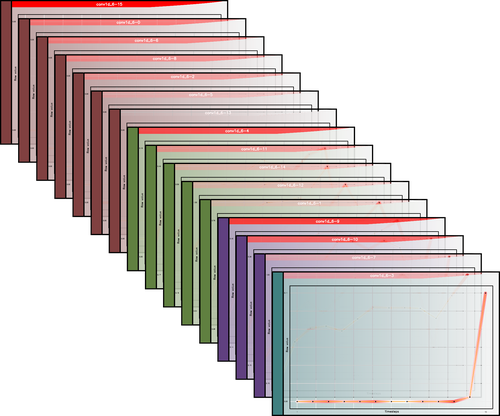}
\label{fig:conv3_imp_cl}}
\caption{Filters of the third convolutional layer of the network trained on the internet traffic dataset}
\label{fig:conv2}
\end{figure*}

Another interesting influence originates from the output, which can be leveraged to compute the filter's influence. 
This information about the filter importance provides hints regarding the filters that were most influential for a particular prediction. Interestingly, many of the filters in the network contribute nothing for a particular prediction. This information is complementary to the information regarding parts of the input that were responsible for a particular prediction.

In order to obtain this influence, we compute the gradient of the output layer $L$ w.r.t. the current layer $l$. This provides us with a point-wise estimate about how each value impacts the output activation $a^{L}$. In this case, again both positive and negative influences are equally important to us. 
As we are interested in the overall influence of a particular filter, therefore, point-wise influence estimates can be reduced to a single value by taking 1p-norm of the influence vector. 
Computing the 1p-norm of the influence vector retains the information encapsulated in the vector, by taking the sum of the absolute influences of each of its components, which provides a good estimate regarding the overall importance of the filter. 
Eq.~\ref{eq:outputInfluence2} provides the mathematical formulation of the influence of layer $l$ on the final output $a^{L}$.

\begin{equation} \label{eq:outputInfluence1}
\delta_{j}^{l} = \frac{\partial a^{L}}{\partial a_{j}^{l}} \frac{\partial a_{j}^{l}}{\partial z_{j}^{l}}
\end{equation}

\begin{equation} \label{eq:outputInfluence2}
    I_{output}^{l} = \sum_{j} \abs{\delta_{j}^{l}}
\end{equation}

Fig.~\ref{fig:conv3_raw} visualizes the third convolutional layer of the network trained on the internet traffic dataset. Fig.~\ref{fig:conv3_imp} enhances the filter view by including the filter importance information computed using Eq.~\ref{eq:outputInfluence2} after applying the min-max normalization along with the input saliency information. 

\begin{proposition}[Zero influence]\label{zeroInfluence}
For extremely confident predictions (with probability of either 0.0 or 1.0) in case of classification, the influence dies off.
\end{proposition}

\begin{proof}
Let $\mathbf{x} \in \mathbb{R}^{D}$ be the input %
and $\mathbf{\hat{y}} \in \mathbb{R}$ ($\mathbf{\hat{y}} \in \mathbb{R}^{C}$ in case of multi-class classification where C denotes the number of classes) be the prediction by the system. For binary classification, the output probability is usually obtained by the application of sigmoid activation, while in the case of multi-class classification, the output probability is obtained by the application of the softmax activation function. These activation function can be considered as the last layer $L$ in the network. 
The gradient of the sigmoid and the softmax activation function w.r.t. to its input is presented in Eq.~\ref{eq:activationSigmoid} and Eq.~\ref{eq:activationSoftmax} respectively.

\begin{equation} \label{eq:activationSigmoid}
	\delta^{L} = \hat{y}(1 - \hat{y})
\end{equation}

\begin{equation} \label{eq:activationSoftmax}
    \delta^{L}_{j} = 
    \begin{dcases}
	    \hat{y_i}(1 - \hat{y_i}), & \text{if i = j} \\ 
	    \hat{y_i}(- \hat{y_j}), & \text{otherwise} \\
	\end{dcases}
\end{equation}

In case of extremely confident predictions, the system makes binary predictions where the probabilty either goes to zero or one i.e. $\hat{y} \in \{0, 1\}$ ($\hat{y} \in \{0, 1\}^C$ in case of multi-class classification where C denotes the number of classes). 
Therefore, during backpropagation, the gradient dies off due to multiplication by zero as highlighted in Eq.~\ref{eq:activationSigmoid} and Eq.~\ref{eq:activationSoftmax} (either the first term or the second term goes to zero due to presence of saturated values). This results in no gradient to previous layers for the computation of the filter influence or saliency for that matter.

\end{proof}

One possible solution to overcome the problem of obtaining no influence values (Proposition~\ref{zeroInfluence}) is to employ temperature-augmented softmax in multi-class classification settings by using $T > 1$ as the temperature (Eq.~\ref{eq:temperaturedSoftmax}). Using $T$ as the temperature will also inversely impact the values of the Jacobian, therefore, should be kept very close to 1.0 just to make sure that extremely confident predictions are avoided.
\begin{equation} \label{eq:temperaturedSoftmax}
	\hat{y_i} = \frac{\exp(a^{L-1}_{i} / T)}{\sum_{j} \exp(a^{L-1}_{j} / T)}
\end{equation}

\input{Clustering}

\subsection{Inverse Optimization}\label{sec:invOptim}

\begin{figure*}[h!]
\centering
\subfloat[Start (8.34\% conf)]{\includegraphics[width=0.44\linewidth]{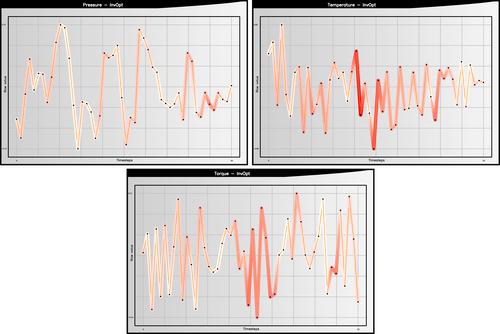}
\label{fig:inv_opt_start_cls}}
\hfil
\hfil
\hfil
\subfloat[End (99.88\% conf)]{\includegraphics[width=0.44\linewidth]{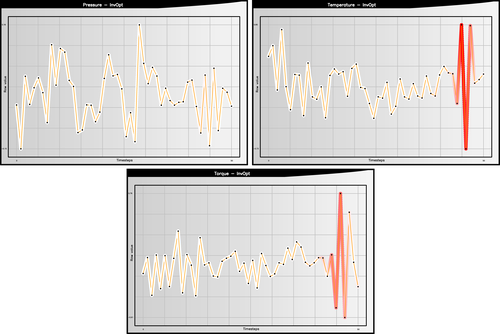}
\label{fig:inv_opt_end_cls}}
\caption{Inverse optimization (classification use-case)}
\label{fig:inv_opt}
\end{figure*}

\begin{figure*}[h!]
\centering
\subfloat[Start (Forecasted value: -0.49923)]{\includegraphics[width=0.44\linewidth]{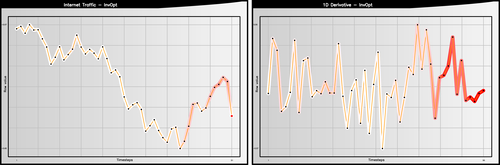}
\label{fig:inv_opt_start_reg}}
\hfil
\hfil
\hfil
\subfloat[End (Forecasted value: -0.95739)]{\includegraphics[width=0.44\linewidth]{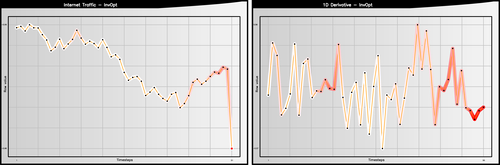}
\label{fig:inv_opt_end_reg}}
\caption{Inverse optimization (regression use-case)}
\label{fig:inv_opt_regression}
\end{figure*}

An understanding of the parts of the input that are considered to be most important factor for a particular prediction is of tremendous value. This information is partly highlighted by computing the input influence (Section~\ref{sec:input_influence}). However, this input influence is not very stable (Section~\ref{sec:stability}). Therefore, we use inverse optimization based framework to discover the main sources of variation learnt by the network in the input space. We randomly start from an input and ask the network to modify the signal so as to achieve a particular output. This optimization based approach gives a true picture of what the network considers to be the main reason for achieving a particular prediction.

During training, we use the loss to minimize the discrepancy between the prediction $\hat{y} \in \mathbb{R}$ and ground-truth $y \in \mathbb{R}$, where both $\mathbf{x} \in \mathbb{R}^{D}$ and $y \in \mathbb{R}$ are fixed ($\mathbf{y} \in \mathbb{R}^{C}, \mathbf{\hat{y}} \in \mathbb{R}^{C}$ in case of multi-class classification where $C$ denotes the number of classes). The optimization problem to discover the optimal parameters of the network $\mathcal{W}^{*}$ can be written as:

\begin{equation} \label{eq:optim1}
\mathcal{W}^{*} = \argmin{\mathcal{W}} \frac{1}{|\mathcal{X}|} \sum_{(\mathbf{x}, y) \in \mathcal{X} \times \mathcal{Y}} \mathcal{L}(\phi(\mathbf{x}; \mathcal{W}), y)
\end{equation}

\noindent where $\phi$ defines the map from the input space $\mathbf{x} \in \mathbb{R}^{D}$ to the label space $\hat{y} \in \mathbb{R}$ ($\mathbf{\hat{y}} \in \mathbb{R}^{C}$ in case of multi-class classification) parameterized by the network weights $\mathcal{W} = {\{W^l,b^l\}}_{l=1}^{L}$. Once the network is trained, i.e. we have the $\mathcal{W}^{*}$, the problem can be inverted to discover the input $\mathbf{\hat{x}} \in \mathbb{R}^{D}$ which produces the same output as a particular time-series. This helps in discovery of the main sources of variation learnt by the network. The following optimization problem can be expressed as Eq.~\ref{eq:optim2}. We solve this problem iteratively using gradient descent as indicated in Eq.~\ref{eq:optim3}.

\begin{equation} \label{eq:optim2}
	\mathbf{\hat{x}^{*}} = \argmin{\mathbf{\hat{x}} \; \in \; \mathbb{R}^{D}} \mathcal{L}(\phi(\mathbf{\hat{x}}; \mathcal{W}^{*}), y)
\end{equation}

\begin{equation} \label{eq:optim3}
	\mathbf{\hat{x}_{t+1}^{*}} = \mathbf{\hat{x}_{t}^{*}} - \alpha \; \frac{\partial \mathcal{L}(\phi(\mathbf{\hat{x}}; \mathcal{W}^{*}), y)}{\partial \mathbf{\hat{x}_{t}^{*}}}
\end{equation}

This optimization problem can be efficiently solved again using the backpropagation algorithm.
We initially sample $\mathbf{\hat{x}^{*}_{0}} \sim \mathcal{N}(0, I)$ from a standard-normal distribution.
Since we initialize the input randomly from a normal distribution, the initial time-series looks distorted. This initialization is visualized in Fig.~\ref{fig:inv_opt_start_cls} for the classification use-case. 
Upon passing this randomly initialized series to the trained anomaly detection model, it marked the sequence as non-anomalous (8.34\% probability for the sequence belonging to the anomalous class).
We then optimized this sequence as mentioned in Eq.~\ref{eq:optim3} using gradient descent. The final sequence obtained after optimizing the input for 1000 iterations ($t=1000$) with $\alpha=0.01$ is visualized in Fig.~\ref{fig:inv_opt_end_cls}. Since the original dataset contained point anomalies, the network introduced point anomalies in the initial time-series to convert the non-anomalous sequence to an anomalous one with a probability of 99.88\%. It is interesting to note that since we never introduced any point anomalies in the pressure signal, during inverse optimization, the network also left the pressure signal intact with only minor changes.
We similarly performed inverse optimization tests on the internet traffic dataset (time-series forecasting). The seemingly unimportant channel containing the first-order derivative based on the saliency map was the main factor that the network nudged in order to obtain the same output as the time-series visualized in Fig.~\ref{fig:inv_opt_start_reg}. The inversely optimized series is visualized in Fig.~\ref{fig:inv_opt_end_reg}. The last time-step in the original signal (internet traffic) indicates the forecasted value.

\subsection{Adversarial Examples}\label{sec:advEx}

\begin{figure*}[h!]
\centering
\subfloat[Input (95.76\% conf)]{\includegraphics[width=0.44\linewidth]{Images/Custom-plots/input/input-comb-sal.png}
\label{fig:adv_start_cls}}
\hfil
\hfil
\hfil
\subfloat[Adversarial output (35.55\% conf)]{\includegraphics[width=0.44\linewidth]{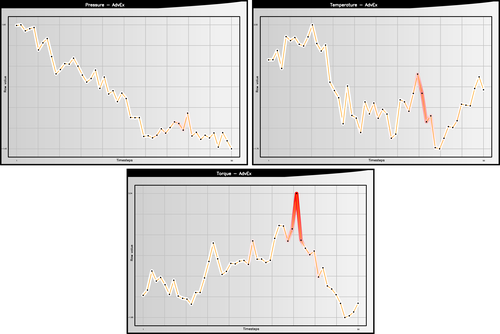}
\label{fig:adv_end_cls}}
\caption{Adversarial examples (classification use-case)}
\label{fig:adv_examples_cls}
\end{figure*}

\begin{figure*}[h!]
\centering
\subfloat[Input (Forecasted value: -0.97946)]{\includegraphics[width=0.44\linewidth]{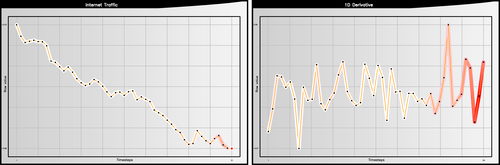}
\label{fig:adv_start_reg}}
\hfil
\hfil
\hfil
\subfloat[Adversarial output (Forecasted value: -0.27947)]{\includegraphics[width=0.44\linewidth]{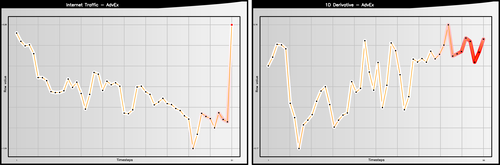}
\label{fig:adv_end_reg}}
\caption{Adversarial examples (regression use-case)}
\label{fig:adv_examples_reg}
\end{figure*}

Deep learning models have been discovered to be highly prone to adversarial noise~\cite{FGSM}. This problem has been very well-studied in prior literature, specifically for image based classification networks~\cite{advLogitPairing}. 
We conducted experiments using iterative FGSM~\cite{iterativeFGSM} (iterative variant of FGSM~\cite{FGSM}) attack on the studied time-series data which is to the best of our knowledge, the first attempt to study these methods for time-series modality. 
We perform these attacks on a regression as well as classification network for time-series. 
This directly provides a hint regarding network's robustness against adversarial noise. In situations where the network is not highly susceptible to adversarial noise, this optimization step can help to answer a more interesting question, which is regarding network's interpretation of the parts of the input that with very minor perturbation can bring maximal change to the output. This highlights network's understanding of the parts of the input that were mainly responsible for particular prediction. 
Having this ability to identify how to modify the input to change the prediction if the model's weren't susceptible to adversarial would have revolutionized design (which was the main intent which led to discovery of these adversarial examples). 
It is important to note that since we perform iterative optimization, this is different than direct saliency computation since the input itself is modified at each time-step.

The FGSM attack performs a single step of optimization to obtain an adversarial example. We denote the adversarial example $\mathbf{x^{adv}} \in \mathbb{R}^{D}$.
The FGSM optimization problem can be represented as:

\begin{equation} \label{eq:adv1}
	\mathbf{x^{adv}} = \mathbf{x} + \epsilon \; sign(\frac{\partial \mathcal{L}(\phi(\mathbf{x}; \mathcal{W}^{*}), y)}{\partial \mathbf{x}})
\end{equation}

The iterative FGSM, instead of solving a single step of optimization, performs iterative refinement of the adversarial noise, therefore, significantly boosting the chances of producing a successful adversarial example. This optimization problem can be written as:

\begin{equation} \label{eq:adv3}
    \mathbf{x^{adv}_{t+1}} = Clip_{\mathbf{x}, \epsilon} \left \{\mathbf{x^{adv}_{t}} + \alpha \; sign(\frac{\partial \mathcal{L}(\phi(\mathbf{x^{adv}_{t}}; \mathcal{W}^{*}), y)}{\partial \mathbf{x^{adv}_{t}}}) \right \}
\end{equation}

\noindent where $Clip_{\mathbf{x}, \epsilon}$ bounds the magnitude of the perturbation to be within $[-\epsilon, \epsilon]$ from the original example $\mathbf{x}$. The value of the original example $\mathbf{x}$ is used to initialize the initial adversarial example $\mathbf{x^{adv}_{0}}$.

Fig.~\ref{fig:adv_start_cls} visualizes an original anomalous sequence present in the anomaly detection dataset. The network successfully marked the sequence as anomalous with a probability of 95.76\%. We then performed the iterative FGSM attack using Eq.~\ref{eq:adv3}, with $\alpha=0.0001$, $\epsilon=0.1$ and $t=1000$. The inverse optimized sequence was predicted to be non-anomalous (35.55\% probability of it being an anomalous sequence) which is visualized in Fig.~\ref{fig:adv_end_cls}.
Consistent with our understanding of the anomaly present in the network, the network reduced the magnitude of the main peak which was mainly responsible for the anomalous prediction. It is interesting to note that the network didn't had such a drastic reduction in the magnitude of the peak so as to achieve such dramatic reduction in the probability of the sequence belonging to the anomalous class. This is indicative of network's susceptibility to adversarial noise.

Adversarial impact on time-series regression task was much more profound. The seemingly non-important first-order turned out to be the main reason for the network's vulnerability. The network mainly altered its prediction due to significant change in the first-order derivative rather than the original signal. Fig.~\ref{fig:adv_examples_reg} highlights this case. We used the same parameters i.e. $\alpha=0.0001$, $\epsilon=0.1$ and $t=1000$. Again, the last time-step in the original signal (internet traffic) indicates the forecasted value.

\begin{figure*}[h!]
\centering
\subfloat[]{\includegraphics[width=0.32\linewidth]{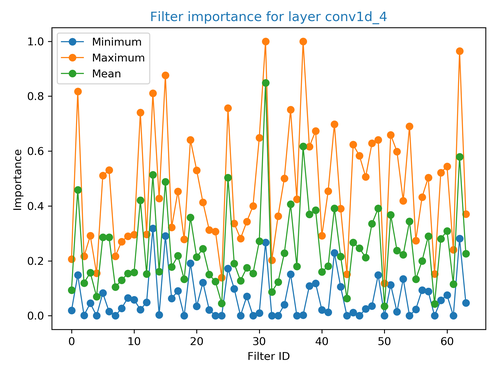}
\label{fig:importance_conv1}}
\hfil
\subfloat[]{\includegraphics[width=0.32\linewidth]{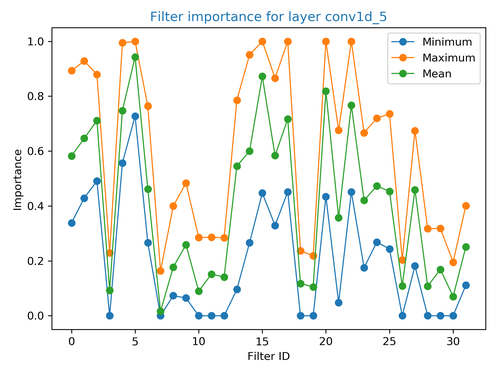}
\label{fig:importance_conv2}}
\hfil
\subfloat[]{\includegraphics[width=0.32\linewidth]{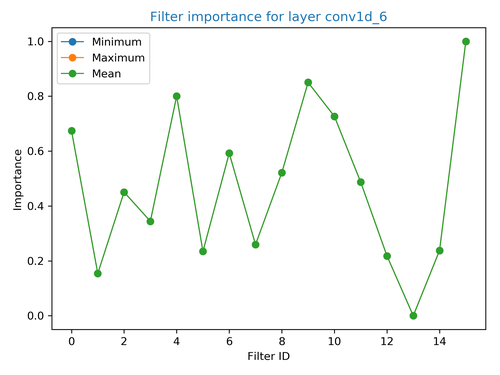}
\label{fig:importance_conv3}}
\caption{Min (a), mean (b) and max (c) filter importance computed over the entire dataset for the first, second and the third convolutional layer on the internet traffic dataset}
\label{fig:filter_importance}
\end{figure*}

\begin{figure*}[h!]
\centering
\subfloat[]{\includegraphics[width=0.32\linewidth]{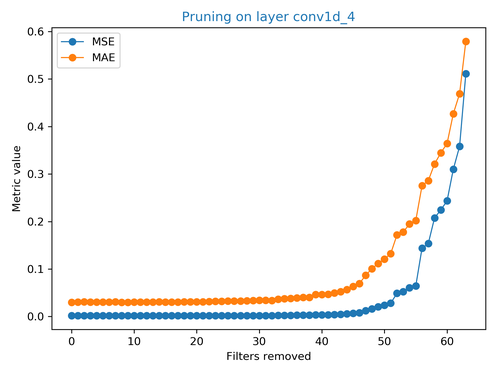}
\label{fig:conv1_prune}}
\hfil
\subfloat[]{\includegraphics[width=0.32\linewidth]{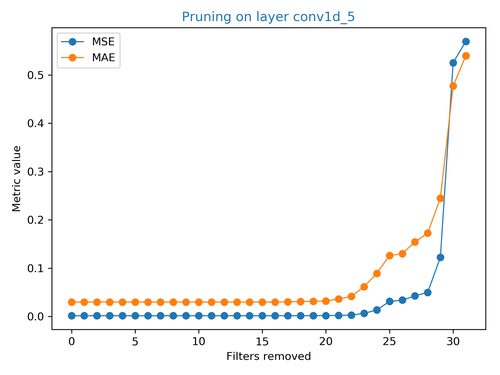}
\label{fig:conv2_prune}}
\hfil
\subfloat[]{\includegraphics[width=0.32\linewidth]{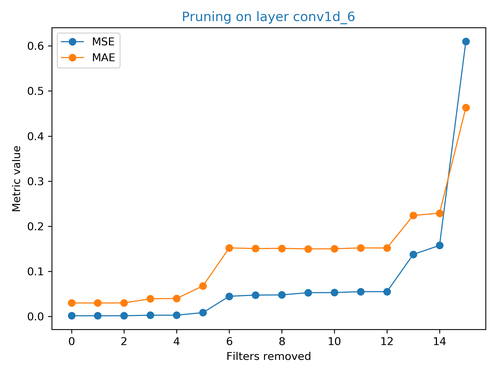}
\label{fig:conv3_prune}}
\caption{Test set performance of the network after pruning the specified number of filters from the first, second and the third convolutional layer on the internet traffic dataset}
\label{fig:pruning}
\end{figure*}

The focus of our work is not exploration of the adversarial examples or guarantees against adversarial robustness, but on an intuitive understanding of their existence and network's susceptibility to it. There are more sophisticated attacks like Carlini and Wagner~\cite{carliniAndWagner} which are extremely effective in exploiting the network. The focus here is to promote interpretability and understanding of the network.

\input{Pruning}

%% file: Clustering.tex
\subsection{Filter Clustering}\label{sec:cluster}

\begin{figure*}[h!]
\centering
\includegraphics[width=\linewidth]{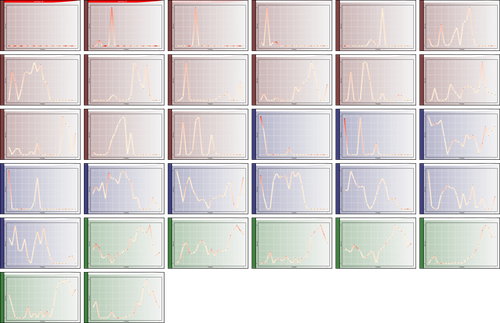}
\caption{Grid view of all the filters equipped with the cluster information (second convolutional layer of the the network trained on the anomaly detection dataset)}
\label{fig:grid_conv_cl}
\end{figure*}

Deep network are great at exploiting redundancy~\cite{deepLearningRedundancy}, therefore, it is important to get a measure of the diversity present in the network. In order to capture this diversity, we perform filter clustering. This clustering phase helps us in discovering the distinct types of filters present in the network as a notion of the diversity it attained during training. We cluster filters based on their activation pattern i.e. filters with similar activating patterns are essentially capturing the same concept. This clustering is also helpful in reducing the information overload for the user in the visualization phase where only the most salient filters from each cluster can be visualized.

As we are only interested in the similarity between the activation pattern rather than the actual magnitude and the shifting of the activation pattern (e.g. invariance to the activation at the start or at the end of the peak), we first align the activations of the different filters in a particular layer. Since we are operating with 1D signals (each filter outputs a 1D activation vector), therefore, in order to compute the similarities between the filters, we leveraged a technique which is very common in time-series analysis community for alignment called as Dynamic Time Warping (DTW)~\cite{fastDTW}.

We encode each filter via its activation vector $\mathbf{a} \in \mathbb{R}^{d}$ where $d$ denotes the dimensionality of the activation. 
The algorithm first creates a distance matrix between the every two activation vectors, $\mathbf{a_m} \in \mathbb{R}^{d}$ and $\mathbf{a_n} \in \mathbb{R}^{d}$. We call the distance matrix as $\mathrm{DTW}$ where $\mathrm{DTW}[i, j]$ gives the distance between the activation vectors $\mathbf{a_m^{1:i}}$ and $\mathbf{a_n^{1:j}}$ with the best alignment. The $\mathrm{DTW}$ matrix can be effectively computed by consistent application of Eq.~\ref{eq:dtw} where the distance metric $D(i, j)$ is the euclidean distance.

\begin{equation} \label{eq:dtw}
	\mathrm{DTW}[i, j] = D(\mathbf{a_m^i}, \mathbf{a_n^j}) + \min 
	(\begin{bmatrix}
    \mathrm{DTW}[i-1, j] \\
    \mathrm{DTW}[i, j-1] \\
    \mathrm{DTW}[i-1, j-1]
    \end{bmatrix})
\end{equation}

\begin{equation} \label{eq:dtw2}
	D(\mathbf{a_m^i}, \mathbf{a_n^j}) = \norm{\mathbf{a_m^i} - \mathbf{a_n^j}}_2
\end{equation}

Once the $\mathrm{DTW}$ matrix is computed, $\mathrm{DTW}[d, d]$ can be used as a measure of the minimum possible distance to align the two activation vectors where $d$ is the dimensionality of the activation vectors. Therefore, we use this distance to cluster the activation vectors together.

When it comes to clustering, K-Means appears to be the most common choice for any problem. However, K-Means operates with euclidean distance as the distance metric. Switching to DTW as the distance metric with K-Means results in either unreliable results or even convergence issues.
Therefore, we performed hierarchical (agglomerative) clustering using DTW where clusters which are closest in terms of distance are combined together to yield a new cluster during every iteration of the algorithm until all the data points are combined into one cluster. There are different possible linkage methods that can be employed to compute the distance between the clusters. In our case, we used the complete linkage to compute the distance between clusters. Complete linkage finds the maximum possible distance $d_{CL} \in \mathbb{R}$ between the two clusters $G$ and $H$ using pairs of points $i$ and $j$ from $G$ and $H$ respectively such that the distance $d_{ij} \in \mathbb{R}$ between the selected points is maximum. This is highlighted in Eq.~\ref{eq:complete_linkage}.

\begin{equation} \label{eq:complete_linkage}
	d_{CL} (G, H) = \max_{i \in G, \; j \in H} d_{ij}
\end{equation}

\begin{figure}[!t]
\centering
\includegraphics[width=\linewidth]{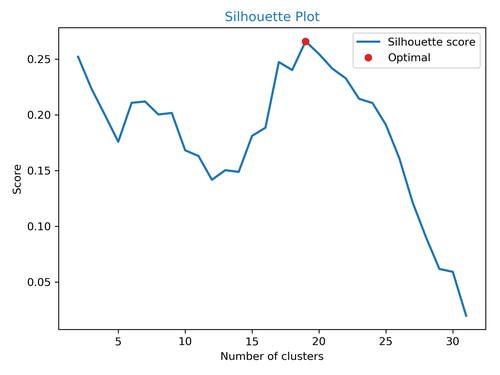}
\caption{Silhouette plot for deciding the initial number of clusters for the second convolutional layer on the anomaly detection dataset}
\label{fig:silhouette_plot}
\end{figure}

The system builds a dense hierarchy of clusters. %
Since we have a dense hierarchy, the user can navigate the hierarchy slicing the y-axis at different points to obtain different number of clusters. Since it is very sensible to start with the best estimation of the possible number of clusters rather than randomly starting in the middle or at the end, we employ the silhouette method to select the initial number of clusters. 
We plot the silhouette scores for each possible number of clusters (except the case where every point is classified into one cluster or where every point is classified into a separate cluster) and then select the initial number of clusters to be at the point where the maximum silhouette score is obtained. This process is illustrated in Fig.~\ref{fig:silhouette_plot}.

Fig.~\ref{fig:conv3_imp_cl} provides a depiction of equipping the filters with the cluster information trained on the internet traffic dataset.
We also tested this clustering strategy on the anomaly detection dataset and visualized the resulting clusters in a grid view for clarity. This visualization is presented in Fig.~\ref{fig:grid_conv_cl}. It can be seen that the silhouette method did a reasonable job in selecting the initial number of clusters.

%% file: Pruning.tex
\subsection{Network Pruning}\label{sec:netPrune}

As a sanity check for the utility of the information contained in the computed influences, we performed pruning of the network based on these computed influences. We use the filter influence and prune filters based on their influence. The filters with least influence are pruned first followed by filters with maximum influence. Since we would like to prune the network based on this information, it is important to average this influence value over the entire training set. Therefore, the final influence w.r.t. the output can be written as:

\begin{equation} \label{eq:pruning1}
I_{output}^{l} = \frac{1}{|\mathcal{X}|} \sum_{\mathbf{x} \in \mathcal{X}} I_{output}^{l}(\mathbf{x})
\end{equation}

We visualize the minimum, maximum as well as mean importance of every filter computed over the entire training set of the internet traffic dataset in Fig.~\ref{fig:filter_importance}. Fig.~\ref{fig:pruning} visualizes the results of pruning based on these influences. We start pruning with the least influential filter until only one filter is left. 
It is important to note that we fine-tune the network for 10 epochs after the pruning step in order to adjust the network weights to compensate for the missing filter.

Table~\ref{tab:faithfulness_eval} provides results regarding faithfulness of the computed influences where we prune the the corresponding most and least influential filter of a particular layer without any fine-tuning. We will discuss this in detail in Section~\ref{sec:faithfulness}. For the sole purpose of pruning to accelerate inference and reduce model size, we refer readers to more sophisticated techniques dedicated to pruning relying on second-order gradient information w.r.t. the loss~\cite{taylorSeriesExpPruning,gazePredFisherPruning}.

%% file: Evaluation.tex
\section{Evaluation} \label{sec:evaluation}

\begin{table*}
\scriptsize
\renewcommand{\arraystretch}{1.0}
\caption{Influence faithfulness test for the internet taffic dataset}
\label{tab:faithfulness_eval}
\centering
\begin{tabular}{|c|c|c|c|c|c|}
\hline
\bfseries Experiment Number & \bfseries Layer & \bfseries Filter Influence & \bfseries MSE & \bfseries MAE & \bfseries Percent Increase (MSE) \\
\hline\hline

0 & Original Network & - & 0.001608 & 0.030010 & 0.000\% \\
1 & 1st Conv & Min & 0.001609 & 0.030020 & 0.049\% \\
2 & 1st Conv & Max & 0.005172 & 0.062662 & 221.608\% \\
3 & 2nd Conv & Min & 0.001608 & 0.030010 & 0.000\% \\
4 & 2nd Conv & Max & 0.007728 & 0.068351 & 380.553\% \\
5 & 3rd Conv & Min & 0.001608 & 0.030010 & 0.000\% \\
6 & 3rd Conv & Max & 0.130290 & 0.247738 & 8002.298\% \\
\hline

\end{tabular}
\end{table*}

The three major desirable properties for any interpretability method are faithfulness, stability and explicitness/intelligiblity~\cite{senn2018}. This section provides an analysis of the TSViz framework on the basis of these three properties.

\subsection{Explicitness/Intelligibility} \label{sec:explicitness}

Explicitness or intelligibility captures the notion of understandability of the explanations provided by the system. Both the input and the output modalities are well-understood by the humans, but the intermediate representations aren't. Therefore, we try to interpret these intermediate representations in terms of their influence on the input and the output. Since both the input as well as the output space are interpretable for humans, this makes the interpretability of the TSViz influence tracing algorithm easy. Adversarial examples and inverse optimization also operate on the input space making them intelligible.

\subsection{Faithfulness} \label{sec:faithfulness}

Faithfulness captures the notion of the reliability of the computed relevance. The true relevance is subjective and can vary from task to task. Influence of noise can be considered as relevant in some cases but might be counter productive to consider in others~\cite{robustnessOfInterpretabilityMethods}. Since we compute the exact influence using gradients and backpropagation, it is a reliable indicator. As a sanity check which is common among literature, filters can be removed from the network to assess their impact on the final performance~\cite{senn2018}.

For the regression network trained on the internet traffic dataset, we removed the most and the least influential filters from the first convolutional layer to assess their impact on the final loss. As per our expectation, pruning the most influential filter had a strong impact on the final performance as compared to pruning the least important filter.

Fig.~\ref{fig:filter_importance} provides a depiction of the filter importance (minimum, maximum and mean importance) computed over the entire training set of the internet traffic dataset. We used this mean importance to remove the most and the least important filter from each of the three convolutional layers on the network. 
Table~\ref{tab:faithfulness_eval} summarizes the results for the faithfulness experiment. 
It is evident from the results that removing the most important filter from a layer had a very significant impact on the performance as compared to pruning the least important filter. Since we directly set the weights of the corresponding filter to zero, therefore, as we ascent the layer hierarchy, the impact of pruning a particular filter was more profound (since it had a direct influence on the result). Pruning also reduces the expected value of the output resulting in a significantly deviated prediction. These results advocate that the computed influence was indeed faithful.

\subsection{Stability} \label{sec:stability}

Since we use the first-order gradient to trace the influence due to it's direct interpretation for humans, this results in unstable explanations due to noise. Interpretability, therefore, sometimes leads to wrong conclusion regarding the smoothness of the decision boundary which is not the case in reality~\cite{FGSM}. Most interpretability methods suffers from this inherent weakness due to reliance on first-order gradients~\cite{robustnessOfInterpretabilityMethods}.
Employing second-order methods can resolve the stability issue, but will make it significantly difficult for the humans to comprehend the gained knowledge.

%% file: TSViz_Tool.tex
\section{Implementation} \label{sec:implementation}

\begin{figure}[t]
\centering
\includegraphics[width=\linewidth]{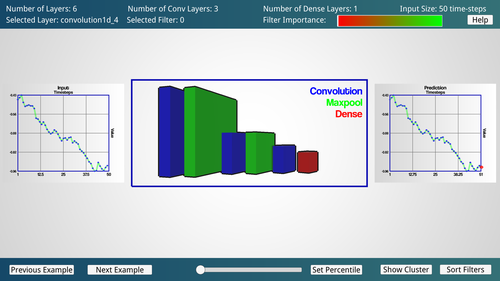}
\caption{Network overview screen (for regression use-case)}
\label{fig:start_screen}
\end{figure}

\begin{figure*}[h]
    \centering
    \includegraphics[width=\textwidth]{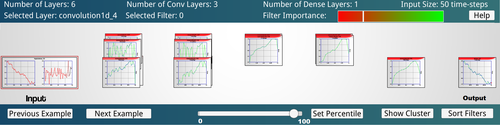}
\caption{Application of the percentile filter on the detailed view (Second level)} 
    \label{fig:percentile_filter}
\end{figure*}

We used Keras with TensorFlow backend~\cite{chollet2015keras,tensorflow2015-whitepaper} developed specifically for deep learning. TensorFlow is an automatic differentiation package which enables automatic computation of the gradients. We leveraged this capability to compute the influences as described in section ~\ref{sec:approach}. 

We developed a novel 3D framework for visualization and demystification of any deep learning model for time-series analysis leveraging the potential of Unity Game Engine~\cite{unityGameEngine}. The user-interface communicates with the back-end which is exposed as a RESTful API. This decouples the model from the visualization aspect. 
Even though the focus of our work is on time-series data, the system is generally applicable to any deep learning model as it is only dependent on the effective computation of the gradients. Our aim is to develop a monolithic framework for time-series deep learning models uncovering all possible demystification aspects. 

The first view in the visualization presents the user with an overview of the network. This gives the user a chance to get acquainted with the model in question. A sample visualization of the first screen is presented in Fig.~\ref{fig:start_screen}. 
The second level provides an overview of most influential/important filters in each layer leveraging the influence computation framework (Section~\ref{sec:influence}). The third view enhances the presented information by clustering the relevant filters together to gain insights regarding the diversity present in the network (Section~\ref{sec:cluster}). The fourth and the fifth level are dedicated for inverse optimization (Section~\ref{sec:invOptim}) and adversarial examples (Section~\ref{sec:advEx}) respectively.

There is usually high interest in visualizing the most important filters from the network since they are indicative of the most important parts of the network leveraged for the prediction. Therefore, we integrated a percentile view where the user the can select the percentile of filters to be viewed based on their importance. This significantly helps in reducing the amount of information presented to the user.
Fig.~\ref{fig:percentile_filter} provides an example application of the percentile filter onto the second level view of the network in the tool. Another possible way to reduce information overload for the user is to visualize the most salient filters from each cluster (Section~\ref{sec:cluster}).

%% file: Discussion.tex
\section{Discussion} \label{sec:discussion}

The visualization enabled a detailed inspection of the network which highlighted many different aspects of the network's learning employed in this study.

\begin{itemize}
\item Most of the filters in the network were useless i.e. they contributed nothing to the final prediction for that particular input. These filters changed as the stimulus to the network changed indicating that some filters were specialized for a particular input.
\item Many of the filters had very similar activating patterns in the network which were assigned to the same filter cluster. This highlighted the aspect of lack of diversity in the trained network. This is consistent with findings of Denil et al. (2013)~\cite{deepLearningRedundancy} where they analyzed the amazing capabilities of deep learning models in exploiting redundancy.
\item Despite of the improvement in performance with the addition of the first-order derivative of the original signal, most of the filters strongly attended to the original signal as compared to the first-order derivative in the time-series forecasting task. On the other hand, when evaluating the adversarial examples and inverse optimization, the network exploited that first-order derivative in order to significantly impact the prediction with only minor modifications in the actual signal.
\item The network mostly focused on the temperature and torque for detecting the anomalies as we never introduced any synthetic anomalies in the pressure signal in the time-series classification task.
\end{itemize}

We argue that there is no perfect way for the interpretability of these models. Therefore, we inspect the model from many different angles in order to come up with a range of different explanations.
We are currently working on extending this work by tracing the influence of particular training examples on the network using influence functions~\cite{understandingDLPredviaInfluenceFunctions}. This will also enable discovery of dataset errors and model debugging. Another area that can greatly enhance the utility of the system is to provide a human understandable description of the intermediate filters by computing its relation to some of the most commonly used operators in the time-series analysis community.

%% file: Conclusion.tex
\section{Conclusion} \label{sec:conclusion}

We proposed a novel framework (TSViz) for interpretability of deep learning based time-series analysis models. The framework enabled an understanding of the model as a parametric function. The different views available within the framework enabled in-depth exploration of the network. This exploration will help in understanding of the network itself as well as enable new improvements within the architecture by the insights gained through uncovering the different aspects of the trained model.